# An attempt to generate new bridge types from latent space of denoising diffusion Implicit model

Hongjun Zhang

Wanshi Antecedence Digital Intelligence Traffic Technology Co., Ltd, Nanjing, 210016, China

583304953@QQ.com

**Abstract:** Use denoising diffusion implicit model for bridge-type innovation. The process of adding noise and denoising to an image can be likened to the process of a corpse rotting and a detective restoring the scene of a victim being killed, to help beginners understand. Through an easy-to-understand algebraic method, derive the function formulas for adding noise and denoising, making it easier for beginners to master the mathematical principles of the model. Using symmetric structured image dataset of three-span beam bridge, arch bridge, cable-stayed bridge and suspension bridge , based on Python programming language, TensorFlow and Keras deep learning platform framework , denoising diffusion implicit model is constructed and trained. From the latent space sampling, new bridge types with asymmetric structures can be generated. Denoising diffusion implicit model can organically combine different structural components on the basis of human original bridge types, and create new bridge types.

## 0  Introduction

Only the six primary generative artificial intelligence algorithms attempted by the author (variational autoencoder, generative adversarial network, pixel convolutional neural network, normalizing flow, energy-based model, denoising diffusion implicit model) have basically met the requirements of bridge-type innovation. Considering the existence of other more advanced algorithms, generative artificial intelligence technology has met the ability standard of virtual assistant for bridge scheme designers. This technology can easily be horizontally extended to fields such as landscape and architectural design. (Note: This only discusses image generation. In fact, generative artificial intelligence technology can also generate text, voice, video, etc., which has great application value in the engineering field, such as the preparation and review of project tender book and general specification.)

The ideas for implementing this technology in the field of bridge design include: ① Artificial intelligence enterprises, universities, entrepreneurial teams, etc., expanding artificial intelligence technology to the field of bridges to improve ecological layout; ② Large design institutes can establish and acquire research and development teams, or collaborate with external parties to develop or outsource research and development tasks. Generally speaking, avoiding basic algorithms and fine-tuning existing pre trained models using specialized bridge datasets is a low-cost and fast technology route to produce results, and product performance can also meet industry requirements. Due to factors such as macroeconomic environment, marketing, human and hardware costs, it is difficult to make profits in the short term. At present, this technology can only provide creative styling, and it is still difficult to generate a complete scheme design, let alone preliminary design and construction drawing design. Therefore, there are still considerable limitations and room for technological improvement.

Recent large-scale diffusion models, such as OpenAI's DALL-E 2 and Google's Imagen, have demonstrated incredible generation capability of text to image. The diffusion model[1-3], as a powerful generative modeling technique, has surpassed the classical generative adversarial network (GAN) in image generation quality, and is easy to train and expand. It has become the preferred choice for developers, especially for text to image applications. Diffusion refers to the process of gradually converting an image into noise. By using a trained diffusion model, it is possible to simulate the reverse operation of diffusion, gradually denoising from noise to generate images.

This article establishes an denoising diffusion implicit model(DDIM) [2,4], using the same bridge image dataset as before[5-9], and further attempts geometric combination innovation of bridge types (open source address of this article's dataset and source code: https://github.com/QQ583304953/Bridge-DDIM).

# 1 Introduction to denoising diffusion implicit model
## 1.1 Overview

The breakthrough paper on diffusion model (denoising diffusion probability model, DDPM) was published in the summer of 2020 [1], and later evolved into denoising diffusion implicit model [2], latent diffusion model [3], etc. Currently, popular AI painting models such as DALL-E 2 and Stable Diffusion are all based on latent diffusion model for their core algorithms.

Given the image dataset, we gradually add a bit of noise. At each step, the image becomes increasingly blurry, with more and more noise components and less information in the original image. Then, we learn a machine learning model that can undo each such step, so we have a model that can generate images from pure random noise.

The denoising diffusion implicit model can be intuitively understood as: the generative model is like a detective, seeing the victim's final decaying body, as long as the time of death is known, it can gradually analyze and restore the scene of the victim's murder, regardless of how many types of erosion the body has experienced. How did this detective acquire this skill? Answer: During his previous studies in university, he conducted many experiments on corpse decay in the laboratory, observing the effects of various erosion methods and durations on the condition of corpses, constantly correcting his own cognition, and ultimately achieving success in his cultivation.

## 1.2 Operation of adding noise to images

If there is an image $X_0$, we hope to gradually erode the image through a series of steps, so that the final erosion result $X_T$ looks similar to standard Gaussian noise. This step is equivalent to a mathematical function $q$, which can add a small amount of Gaussian noise to an image $X_{t-1}$ to generate a new image $X_t$. If we keep using this function $q$, we will generate a series of progressive noise images ($X_0$, $X_1$, ···, $X_{t-1}$, $X_t$, ···, $X_{T-1}$, $X_T$) as shown in the following figure.

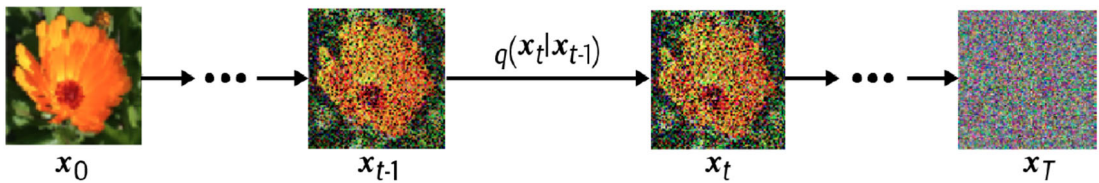

Fig.1 Schematic diagram of adding noise to image

For ease of calculation, the dataset is first subjected to data normalization preprocessing, where image $X_0$ follows a standard normal distribution. (Note: This step is not about adding noise, but about lossless image format conversion.)

1. To ensure that each noisy image follows a standard normal distribution, we need to construct the function $q$ as follows:

$$X_t = \sqrt{1-\beta_t}X_{t-1} + \sqrt{\beta_t}\epsilon_{t-1} \tag{1}$$

In the formula: $\epsilon_{t-1}$ is noise, following a standard normal distribution; $\sqrt{1-\beta_t}$ is the proportion coefficient (signal_rates) of image $X_{t-1}$ information; $\sqrt{\beta_t}$ is the proportion coefficient (noise_rates) of noise information.

$X_{t-1}$ and $\epsilon_{t-1}$ are independent of each other and all follow a standard normal distribution. $(1-\beta_t)$ is the variance of $\sqrt{1-\beta_t}X_{t-1}$, and $\beta_t$ is the variance of $\sqrt{\beta_t}\epsilon_{t-1}$. Because of $(1-\beta_t)+\beta_t=1$, then $X_t$ also follows a standard normal distribution.

In formula (1), $\epsilon_{t-1}$ is randomly selected, so there are countless possibilities for the results of each step.

2. From an image $X_0$ to $X_t$, only one operation of function $q$ is required, without the need for t times operations of function $q$. The reasoning is as follows:

Order $\alpha_t = 1 - \beta_t$, then formula (1) becomes:

$$X_t = \sqrt{\alpha_t}X_{t-1} + \sqrt{1-\alpha_t}\epsilon_{t-1} \tag{2}$$

$$= \sqrt{\alpha_t}\left(\sqrt{\alpha_{t-1}}X_{t-2} + \sqrt{1-\alpha_{t-1}}\epsilon_{t-2}\right) + \sqrt{1-\alpha_t}\epsilon_{t-1} \tag{3}$$

$$= \sqrt{\alpha_t\alpha_{t-1}}X_{t-2} + \sqrt{\alpha_t(1-\alpha_{t-1})}\epsilon_{t-2} + \sqrt{1-\alpha_t}\epsilon_{t-1} \tag{4}$$

$\epsilon_{t-2}$ and $\epsilon_{t-1}$ are independent of each other and all follow a standard normal distribution. Because of $\alpha_t(1-\alpha_{t-1}) + (1-\alpha_t) = 1 - \alpha_t\alpha_{t-1}$, for the operation of adding noise, the following equation meets the requirements:

$$\sqrt{\alpha_t(1-\alpha_{t-1})}\epsilon_{t-2} + \sqrt{1-\alpha_t}\epsilon_{t-1} = \sqrt{1-\alpha_t\alpha_{t-1}}\epsilon' \tag{5}$$

Then:

$$X_t = \sqrt{\alpha_t\alpha_{t-1}}X_{t-2} + \sqrt{1-\alpha_t\alpha_{t-1}}\epsilon' \tag{6}$$

Order $\bar{\alpha} = \alpha_t\alpha_{t-1}\ldots\alpha_1$, it can be obtained by mathematical induction:

$$X_t = \sqrt{\bar{\alpha}}X_0 + \sqrt{1-\bar{\alpha}}\epsilon \tag{7}$$

In the formula: $\epsilon$ is noise, following a standard normal distribution; $\sqrt{\bar{\alpha}}$ is the proportion coefficient (signal_rates) of image $X_0$ information; $\sqrt{1-\bar{\alpha}}$ is the proportion coefficient (noise_rates) of noise information.

3. Using formula (7), convert the dataset into noisy images, as shown in the following figure (using a cosine diffusion schedule with an offset term):

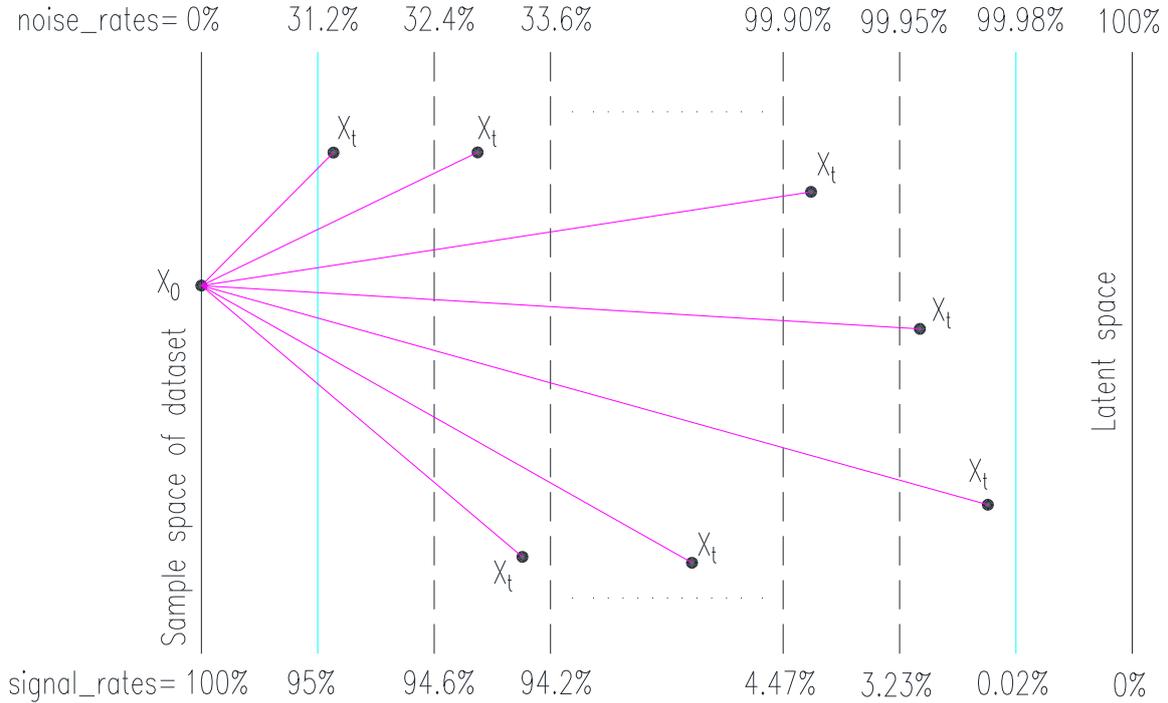

Fig.2 Schematic diagram of converting dataset into noisy images

In the above figure, the proportion coefficient (signal_rates) of image $X_0$ information is set the value ranges from 95% to 0.02%, and the proportion coefficient (noise_rates) of noise information is set the value ranges from 31.2% to 99.98%. The latent space shown in the figure is artificially assumed, corresponding to 100% noise, and is the theoretical result of $X_0$ undergoing infinite times of function $q$ operations. In fact, it is impossible to appear according to the diffusion schedule shown in the figure.

4. Each magenta line segment in Figure 2 represents a single operation of adding noise that uses

formula (7). In general, there is no collinear scenario in Figure 2.

However, it is also possible to artificially construct collinear scenarios using the following methods:

$$X_{20} = \sqrt{\bar{\alpha}_{20}}X_0 + \sqrt{1-\bar{\alpha}_{20}}\epsilon_{20} \tag{8}$$
$$X_{30} = \sqrt{\bar{\alpha}_{30}}X_0 + \sqrt{1-\bar{\alpha}_{30}}\epsilon_{30} \tag{9}$$

$\epsilon_{20}$ and $\epsilon'$ are independent of each other and all follow a standard normal distribution. Because of $(1-\bar{\alpha}_{30}-\sigma^2)+\sigma^2=1-\bar{\alpha}_{30}$, for the operation of adding noise, the following equation meets the requirements:

$$\sqrt{1-\bar{\alpha}_{30}-\sigma^2}\epsilon_{20} + \sigma\epsilon' = \sqrt{1-\bar{\alpha}_{30}}\epsilon_{30} \tag{10}$$

Simultaneous formula (9) and (10):

$$X_{30} = \sqrt{\bar{\alpha}_{30}}X_0 + \sqrt{1-\bar{\alpha}_{30}-\sigma^2}\epsilon_{20} + \sigma\epsilon' \tag{11}$$

Here $\sigma$ can take any value. When $\sigma=0$:

$$X_{30} = \sqrt{\bar{\alpha}_{30}}X_0 + \sqrt{1-\bar{\alpha}_{30}}\epsilon_{20} \tag{12}$$

Observing formula (8) and (12), two different operations using the same noise value, which means that $X_{20}$ is a point on the path of $X_{30}$. Therefore, the two operations of $X_{20}$ and $X_{30}$ are collinear.

In formula (11), when $\sigma \neq 0$, it indicates that the two operations $X_{20}$ and $X_{30}$ are not collinear, which means that the noise values are not the same.

## 1.3 Training neural network to predict noise

In formula (7), there are a total of 4 variables, and when adding noise, it is known that $\bar{\alpha}$、$X_0$、$\epsilon$ values, and we can solve $X_t$ through algebraic operation.

If only knowing $\bar{\alpha}$、$X_t$ values, $\epsilon$ 、$X_0$ cannot be solved through algebraic operations.

However, image features have characteristics such as translation invariance and spatial hierarchy, and only using $\bar{\alpha}$、$X_t$ variables, after training, the neural network can predict $\epsilon$ 、$X_0$. For example, if the left side of a corpse is corroded by an unknown chemical substance, a detective can use the symmetry of the human body to restore it.

1. U-Net neural network to predict noise

In a similar manner to a variational autoencoder, a U-Net consists of two halves: the downsampling half, where input images are compressed spatially but expanded channel-wise, and the upsampling half, where representations are expanded spatially while the number of channels is reduced. However, unlike in a VAE, there are also skip connections between equivalent spatially shaped layers in the upsampling and downsampling parts of the network. A VAE is sequential; data flows through the network from input to output, one layer after another. A U-Net is different, because the skip connections allow information to shortcut parts of the network and flow through to later layers.

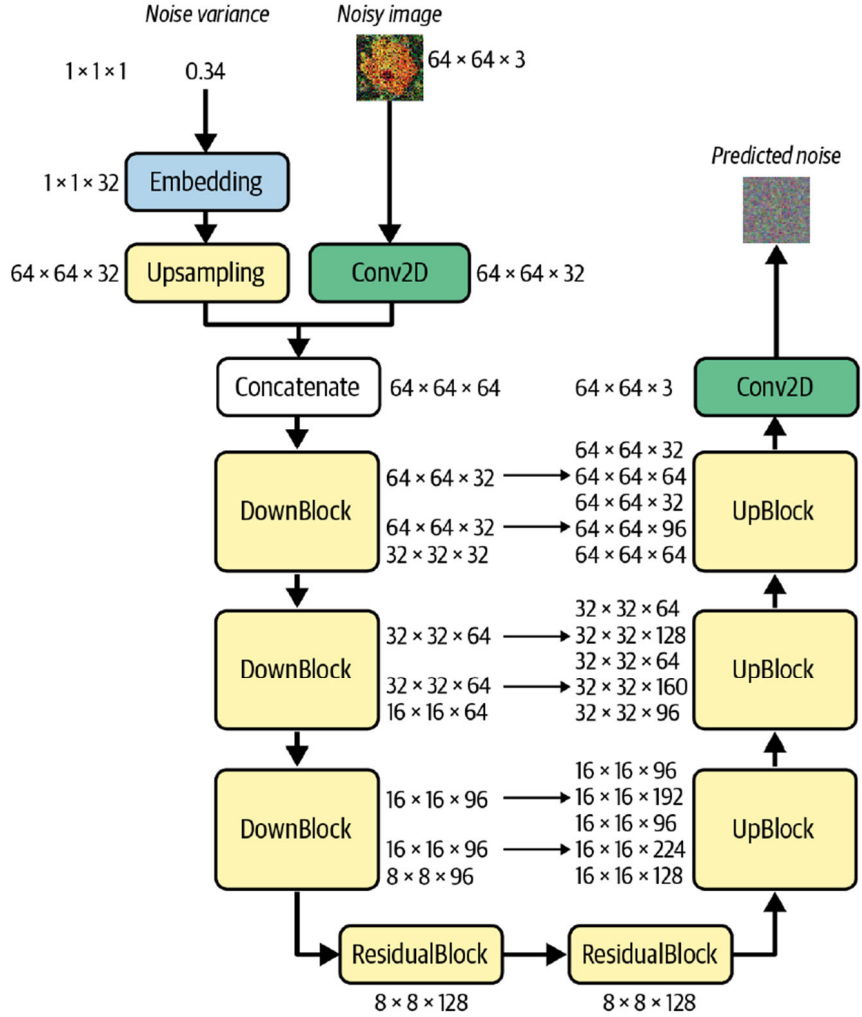

Fig.3 U-net architecture diagram

In formula (7), $1-\bar{\alpha}$ is the variance of $\sqrt{1-\bar{\alpha}}\epsilon$, also known as the square of noise_rates.

The function of U-Net neural network is to input variance, noise images, and calculate the predicted value of noise $\epsilon$. The predicted value of $X_0$ can be obtained by using formula (7).

The neural network is trained by taking the absolute difference between the real noise $\epsilon$ and the predicted noise $\epsilon$ as a loss function.

## 1.4 The operation of denoising to generate images

Starting from the random noise $X_t$ in the latent space, we use a model to gradually denoise until we obtain a meaningful image $X_0$.

The model gets the predicted values of noise $\epsilon$ and $X_0$ in one step. It can be imagined that it is difficult to generate a meaningful image in one step, and the model performance cannot achieve this ability. After all, the latent space sampling points are 100% pure noise, and the maximum noise during previous training was only 99.98%.

Therefore, it is necessary to gradually denoise. At first generate $X_{t-1}$, and then gradually generate the final image through circular iteration.

A collinear scenario is artificially constructed, and $X_{t-1}$ is calculated from $\epsilon$ and $X_0$ predicted by the model. The process is as follows:

$$X_{t-1} = \sqrt{\bar{\alpha}_{t-1}}X_0 + \sqrt{1-\bar{\alpha}_{t-1}}\epsilon_{t-1} \tag{13}$$

$\epsilon_t$ and $\epsilon$ are independent of each other and all follow a standard normal distribution. Because of $(1-\bar{\alpha}_{t-1}-\sigma_t^2)+\sigma_t^2 = 1-\bar{\alpha}_{t-1}$, for the operation of adding noise, the following equation meets the requirements:

$$\sqrt{1-\bar{\alpha}_{t-1}-\sigma_t^2}\epsilon_t + \sigma_t\epsilon = \sqrt{1-\bar{\alpha}_{t-1}}\epsilon_{t-1} \tag{14}$$

Simultaneous formula (13) and (14) :

$$X_{t-1} = \sqrt{\bar{\alpha}_{t-1}}X_0 + \sqrt{1-\bar{\alpha}_{t-1}-\sigma_t^2}\epsilon_t + \sigma_t\epsilon \tag{15}$$

Here $\sigma_t$ can take any value. When $\sigma_t=0$:

$$X_{t-1} = \sqrt{\bar{\alpha}_{t-1}}X_0 + \sqrt{1-\bar{\alpha}_{t-1}}\epsilon_t \tag{16}$$

Therefore, formula (16) realizes $\epsilon_t$ and $X_0$ predicted by the model, and calculates $X_{t-1}$. Then we do the same thing, and we generate $X_{t-2}$, all the way to $X_0$. The larger the value of step t, the higher the $X_0$ precision in the generated image approximation theory.

The process of iterative image generation in a 20-step cycle is shown below:

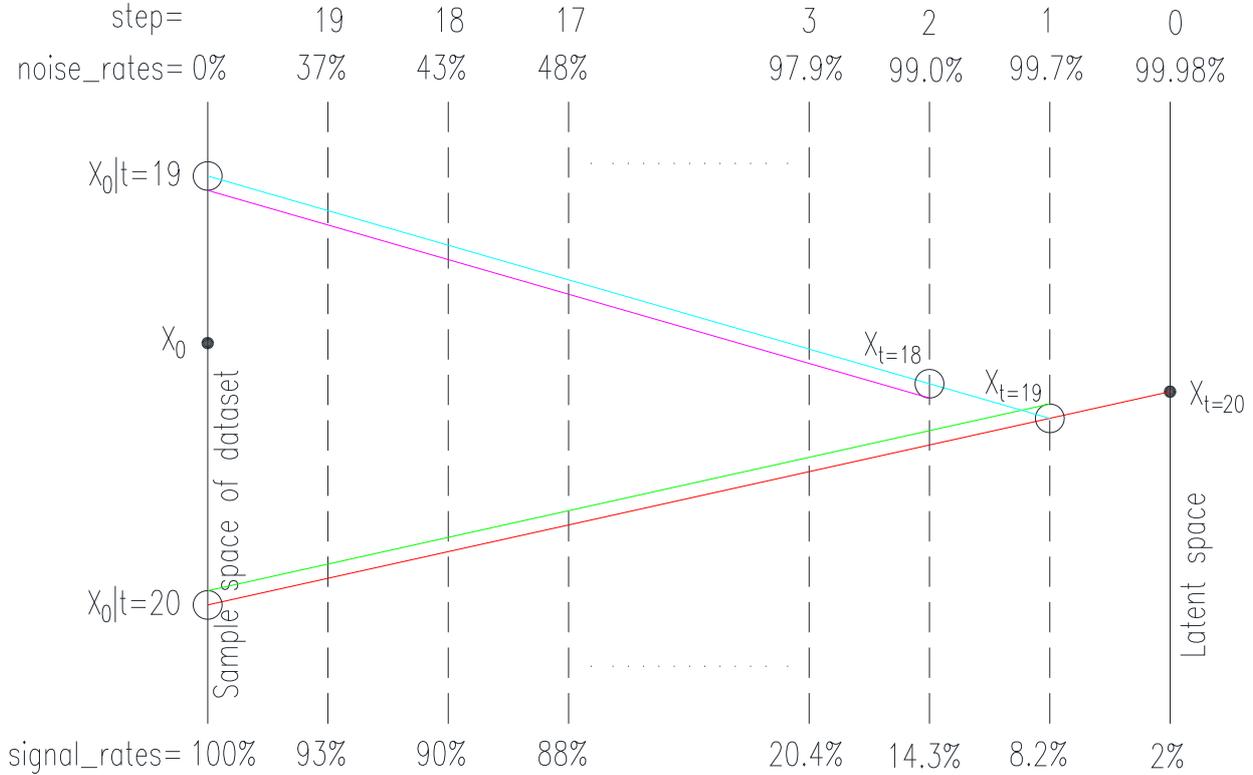

Fig.4 Schematic diagram of generating images by removing noise from latent space

In formula (15), $\sigma_t$ is artificially set to 0. This sacrifices the diversity of the generated image for each sampling point. And this property is exactly what we want, because we want the latent space sampling point to be a definite mapping relationship with the pixel space.

## 2 An attempt to generate new bridge types from latent space of denoising diffusion implicit model

### 2.1 Dataset

Using the dataset from the author's previous paper [5-9], which includes two subcategories for each type of bridge (namely equal cross-section beam bridge, V-shaped pier rigid frame beam bridge, top-bearing arch bridge, bottom-bearing arch bridge, harp cable-stayed bridge, fan cable-stayed bridge, vertical_sling suspension bridge, and diagonal_sling suspension bridge), and all are three spans (beam bridge is 80+140+80m, while other bridge types are 67+166+67m), and are structurally symmetrical.

To be able to train the model quickly, the image size is reduced from 512x128 to 192x48 pixels (the disadvantage is that the clarity is much reduced).

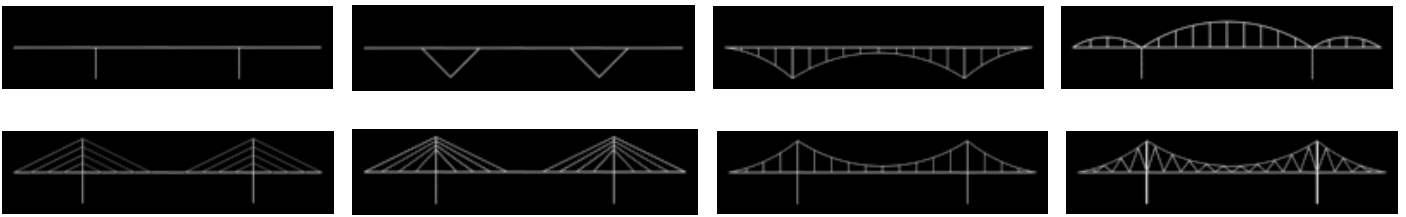

Fig.5 Grayscale image of each bridge facade

Each sub bridge type obtained 1200 different images, resulting in a total of 9600 images in the entire dataset.

## 2.2 Construction and training of model

Based on the Python3.10 programming language, TensorFlow2.10, and Keras2.10 deep learning platform framework, construct and train DDIM[4].

1. Four components of U-Net neural network

(1) the sinusoidal embedding of the noise variance

It can convert a scalar value (the noise variance) into a distinct higher-dimensional vector that is able to provide a more complex representation, for use downstream in the network.

(2) the ResidualBlock

A residual block is a group of layers that contains a skip connection that adds the input to the output. Residual blocks help us to build deeper networks that can learn more complex patterns without suffering as greatly from vanishing gradient and degradation problems.

(3) the DownBlock

Each successive DownBlock increases the number of channels via block_depth (=2 in our example) ResidualBlocks, while also applying a final AveragePooling2D layer in order to halve the size of the image. Each ResidualBlock is added to a list for use later by the UpBlock layers as skip connections across the U-Net.

(4) the UpBlock

An UpBlock first applies an UpSampling2D layer that doubles the size of the image, through bilinear interpolation. Each successive UpBlock decreases the number of channels via block_depth (=2) ResidualBlocks, while also concatenating the outputs from the DownBlocks through skip connections across the U-Net.

2. Training

The loss curve of the training process is shown in the following figure:

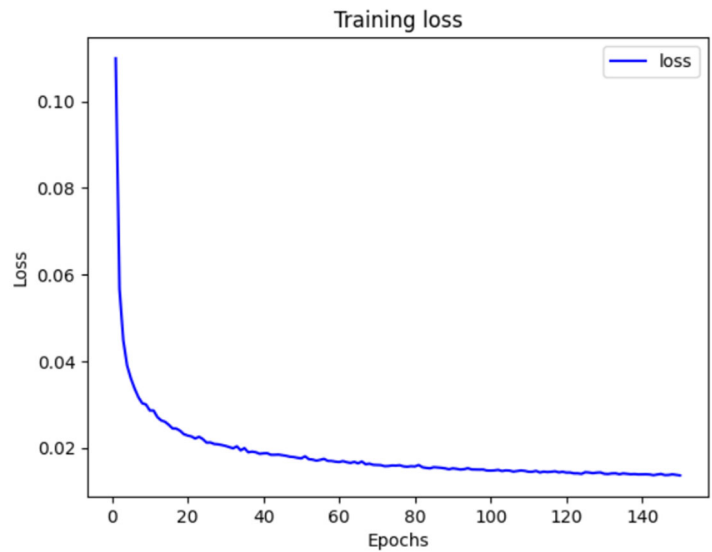

Fig.6 Training loss curve

## 2.3 Exploring new bridge types through latent space sampling

Random sampling from the latent space. Based on the thinking of engineering structure, five technically feasible new bridge types are obtained through manual screening, which were completely different from the dataset:

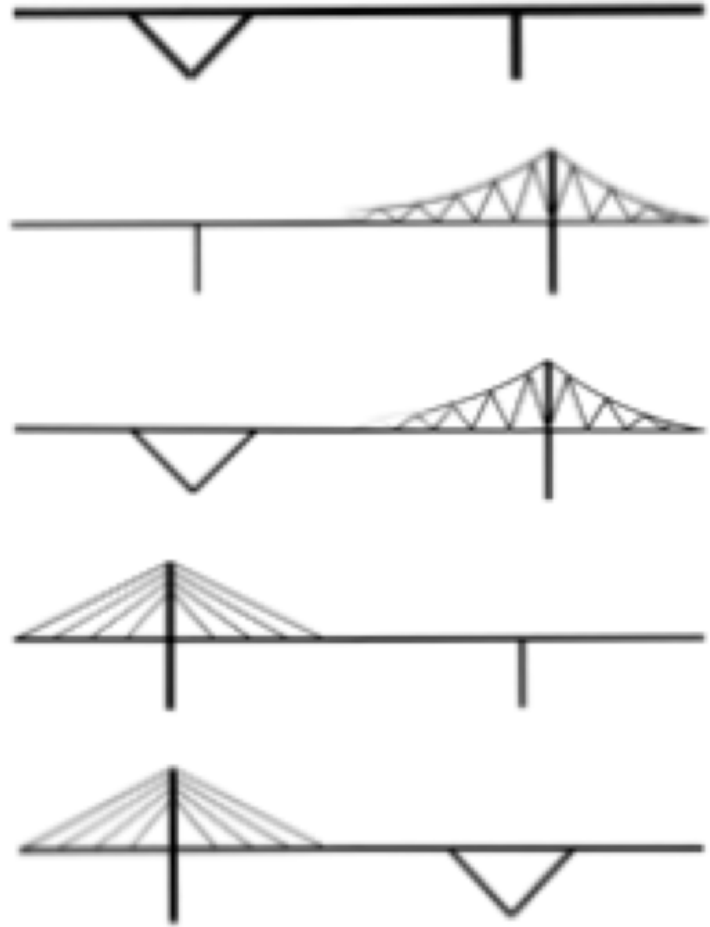

Fig.7 Five new bridge types with feasible technology

The new bridge type here refers to a type of bridge that has never appeared in the dataset, but is created by neural network based on algorithms, which represents the model's innovative ability.

## 3 Conclusion

The performance of the denoising diffusion implicit model is similar to that of generative adversarial network and normalizing flow, and is more creative than variational autoencoder and energy-based model. It can organically combine different structural components on the basis of human original bridge types, creating new bridge types. It has a certain degree of human original ability, which can open up the space of imagination and provide inspiration to humans.